\documentclass[letterpaper, 10 pt, conference]{ieeeconf} 
\IEEEoverridecommandlockouts                     
\usepackage[utf8]{inputenc}
\usepackage[english]{babel}
\usepackage{color}
\usepackage[table]{xcolor}
\usepackage{gensymb}
\usepackage{multirow}
\usepackage{float}
\usepackage{graphics}
\usepackage{graphicx}
\usepackage{caption}
\usepackage{multicol}
\usepackage{subcaption}
\usepackage{color, colortbl}
\usepackage{amsmath}
\usepackage{array}
\usepackage{adjustbox}
\title{\LARGE \bf Adapting Semantic Segmentation Models for Changes \\ in Illumination and Camera Perspective}

\author{Wei Zhou, Alex Zyner, Stewart Worrall, and Eduardo Nebot 
\thanks{W. Zhou, A. Zyner, S. Worrall, and E. Nebot are with the 
Australian Centre for Field Robotics (ACFR) at the University of Sydney, NSW, Australia. E-mails: {\tt \{w.zhou, a.zyner, s.worrall,
e.nebot\}@acfr.usyd.edu.au}.}
}

\begin{document}

\maketitle
\thispagestyle{empty}
\pagestyle{empty}

\begin{abstract}

Semantic segmentation using deep neural networks has been widely explored to generate high-level contextual information for autonomous vehicles. To acquire a complete $180^\circ$ semantic understanding of the forward surroundings, we propose to stitch semantic images from multiple cameras with varying orientations. However, previously trained semantic segmentation models showed unacceptable performance after significant changes to the camera orientations and the lighting conditions. To avoid time-consuming hand labeling, we explore and evaluate the use of data augmentation techniques, specifically skew and gamma correction, from a practical real-world standpoint to extend the existing model and provide more robust performance. 
The presented experimental results have shown significant improvements with varying illumination and camera perspective changes.

\end{abstract}

\section{Introduction}

Autonomous driving has been the subject of significant research and development in recent years. It requires vehicle sensing and processing that can produce a higher level understanding of the urban road environment in order to make driving decisions in a safe manner without input from human drivers.

Semantic segmentation is a vision task to assign class labels to every point in a given image, with the labels providing a more human-like understanding of a scene. In our previous work~\cite{itsc_work}, we successfully implemented real-time semantic segmentation by transferring a pre-trained convolutional neural network (CNN) to our local environment in the surrounds of the University of Sydney (USYD). A local dataset was collected to fine-tune the pre-trained model using a forward facing Point Grey camera covering $56^\circ$ field-of-view (FOV). 

This paper looks at expanding the FOV of our previous work as perception incorporating a wide FOV is crucial to obtain complete information for situation awareness. Fish-eye or omni-directional cameras have been employed to achieve a large FOV, though objects viewed from these sensors appear to be highly distorted and are represented by only a relatively small number of pixels even at a moderate distance. Therefore, we propose to generate a high resolution, wide FOV image by combining a number of pinhole cameras.

\begin{figure}[t]
\begin{center}
\includegraphics[height=4.5cm,width=0.98\columnwidth]{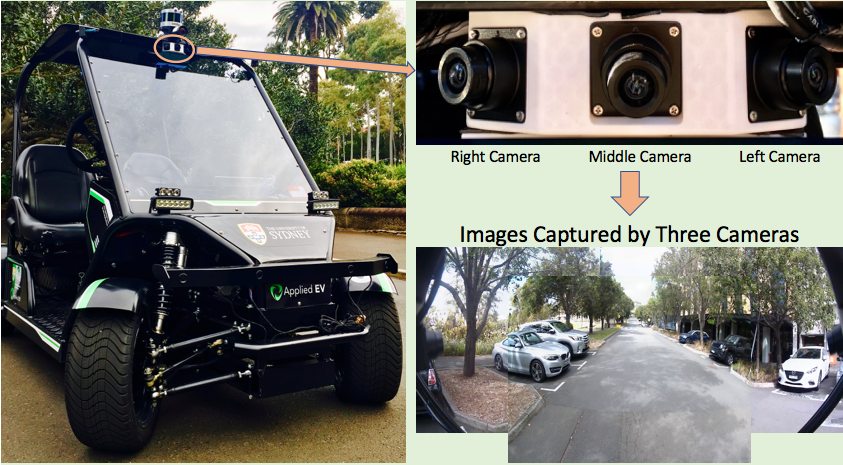}
\caption{\small Autonomous electrical vehicle. There are three NVIDIA GMSL cameras mounted in the front of the vehicle. Images captured by these three cameras can cover a $180^\circ$ field of view. Two side cameras have around $20^\circ$ overlap. The middle camera is tilted $15^\circ$ downwards.}
\label{fig:car}
\end{center}
\end{figure}

We replaced the Point Grey $56^\circ$ camera from our previous work~\cite{itsc_work} with NVIDIA GMSL $100^\circ$ cameras. By mounting three cameras in an array (Fig.~\ref{fig:car}), we are able to cover a $180^\circ$ FOV in front of the vehicle and acquire an acceptable image distortion with low cost. Incorporating multiple overlapping cameras results in redundant information, and to generate a complete picture it is necessary to find the corresponding relationships between the cameras. To achieve this, we propose to stitch together the images from the camera array to provide a single view for processing and visualization. 

Given a wider $180^\circ$ FOV, we found that the previous semantic segmentation models~\cite{itsc_work} trained on forward-view cameras did not show satisfactory results on side-view cameras due to perspective changes. 
In addition, the images in the previous dataset were mostly collected on cloudy days and the model trained from this dataset performed poorly on very sunny days in summer due to the strong shadows and over-exposure. Therefore, further training is necessary to adapt to different perspectives and illumination conditions.

To extend and generalize a CNN model, a significant number of pixel-level annotated images are required which is labor-intensive and time-consuming to generate. The widely used Cityscapes dataset~\cite{cityscapes} required more than one hour to annotate a single image. To minimize the requirement for additional hand labeling of images, data augmentation techniques have been widely employed to artificially expand existing datasets. 

However, most existing works like \cite{autoaugment} and \cite{augmentation_jesus} are applying data augmentations to achieve theoretical improvements on public datasets~\cite{cityscapes,kitti}, whilst it is more practically meaningful to examine specific data augmentation techniques to enhance model robustness with lighting and perspective changes.
For this purpose, we propose to augment the dataset with skewed and gamma corrected images to simulate side-view objects, dark shadows and over-exposure under real-world scenarios.



\section{Related Work}
\label{sec:related work}

This section presents an overview of recent work related to the proposed contributions of this paper. 

\subsection{Semantic Segmentation}
\subsubsection{High-accuracy Networks}
Semantic segmentation has been the subject of significant research, and there are two main techniques used in this area. The first is the encoder-decoder architecture, which have been employed to downsample and upsample images in order to obtain the segmentation results~\cite{fcn,refinenet}. The other technique makes use of atrous convolutions to exponentially expand the receptive fields without losing resolution, and pyramid pooling which enables the network to concatenate features from different levels. \cite{deeplabv2,pspnet,deeplabv3} all employ this strategy and are currently the state-of-the-art when measured against Cityscapes benchmarks~\cite{cityscapes}.

\subsubsection{Real-time Implementation}
In contrast to the significant development towards high-accuracy semantic segmentation, real-time implementation of semantic segmentation has not been a major focus. SegNet~\cite{segnet} was the first network to process images in close to real-time. This is achieved by only saving the pooling indices during encoding. ENet~\cite{enet} uses an encoder-decoder architecture and atrous convolution to significantly reduce the processing time. The image cascade network (ICNet) integrates multiple resolutions and fusion units to further boost the accuracy and can achieve an average of 30 frames per second on $1024 \times 2048$ images.

\subsubsection{Fish-eye Semantic Segmentation}

Omnidirectional vision systems are being increasingly incorporated into autonomous vehicle sensor platforms. Deng et al. \cite{fisheye1} propose an Overlapping Pyramid Pooling Network (OPP-net) for fish-eye semantic segmentation. Varga et al. \cite{supersensor} present several image unwarping strategies for fish-eye cameras and builds a $360^\circ$ perception system. Common issues with utilizing this type of sensor are insufficient image resolutions and large object distortions. In addition, there are few annotated datasets for fish-eye cameras to facilitate the training of a semantic segmentation network. 

\subsection{Generic Data Augmentation}

Supervised Deep Neural Networks (DNNs) are widely known as data-hungry algorithms leading to the requirement for enormous datasets during the training process. The issue is that obtaining labeled data is challenging as it is very time consuming to hand label images for every potential scenario. ImageNet~\cite{imagenet} applied cropping and flipping of the training images to increase the dataset by a factor of 2048. They also alter the intensities for RGB channels to further expand the dataset.
Carlson et al.~\cite{alexa} addressed data augmentation from sensor effect perspectives with chromatic aberration, blur, exposure, noise and color cast on both real and synthetic images. They also proved that a synthetic dataset has to be significantly larger than a real dataset to achieve similar performance. AutoAugment~\cite{autoaugment} used a search algorithm with reinforcement learning to automatically learn the best augmentation policies for different datasets. In general, data augmentation has been widely implemented for the expansion of datasets in order to teach models to be invariant in the data domain and to avoid overfitting~\cite{augmentation_jesus,aug1,aug2}. Few of these approaches however, examine data augmentation from a practical real-world perspective.

\begin{figure}[t]
\begin{center}
\includegraphics[height=1.8cm, width=0.98\columnwidth]{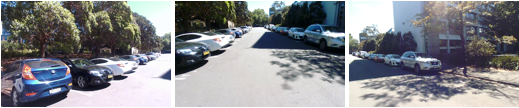}
\caption{\small Left, front and right camera images captured by NVIDIA GMSL cameras.}
\label{fig:nvidia}
\end{center}
\end{figure}

\section{Methodology}
\label{sec:methodology}

In this section, we present the description of our semantic segmentation dataset, the algorithms to generate augmented data for different perspectives and illumination conditions, and the method to stitch multiple camera images.

\subsection{Image Data Preparation}

We collect our local data using an electrical vehicle (EV) shown in Fig. \ref{fig:car}. There are 3 NVIDIA SF3322 GMSL cameras mounted in the front of the EV covering a $180^\circ$ FOV. Each camera has $100^\circ$ horizontal FOV and $60^\circ$ vertical FOV. The front camera is tilted $15^\circ$ downwards and two side cameras are tilted $40^\circ$ to the left and to the right respectively. A NVIDIA DRIVE PX2 is also equipped to provide real-time processing capabilities.

There are several datasets used in this paper to train and evaluate data augmentation performance for semantic segmentation:

\subsubsection{Existing Dataset}

In our previous work~\cite{itsc_work}, we collected image data in the surrounds of the USYD campus using a Point Grey camera in mostly cloudy weather conditions. There were around 170 images with 12 classes manually annotated using the on-line annotation tool LabelMe~\cite{labelme}. The annotation process started with objects in the background, then moved to the foreground, i.e. from `Sky' to `Pedestrian'. This order ensured objects that are distant from the camera can be overruled by objects closer to the camera. We named this dataset as \texttt{USYD\char`_Cloudy\char`_Set}, and applied some image transformations such as flipping, center-cropping, adding noise and blurring to artificially expand the dataset. Models fine-tuned using this dataset were demonstrated to produce good results for our local environment~\cite{itsc_work}. 

\subsubsection{Supplemental Evaluation Datasets}

Due to the change in camera type and a different sensor layout, we annotated 40 images for each of the new NVIDIA GMSL cameras designating them \texttt{USYD\char`_Front\char`_Set}, \texttt{USYD\char`_Left\char`_Set} and \texttt{USYD\char`_Right\char`_Set} respectively. 
Images in these sets have generally challenging conditions such as over-exposure or with harsh shadows (sample images in Fig.~\ref{fig:nvidia}).  In this paper, these three datasets will only be used to validate the performance of transferring models trained on the \texttt{USYD\char`_Cloudy\char`_Set} to new illumination conditions (like shadows) and to different camera perspectives using the specified data augmentations. 

In addition to the local data, we also employ images collected from Brisbane Australia~\cite{day_data,shadow_data} and images used for localization~\cite{cmu} to qualitatively cross-test the semantic segmentation models.

\subsection{Extending Data Augmentation}

\begin{figure}[t]
\begin{center}
\includegraphics[width=0.88\columnwidth]{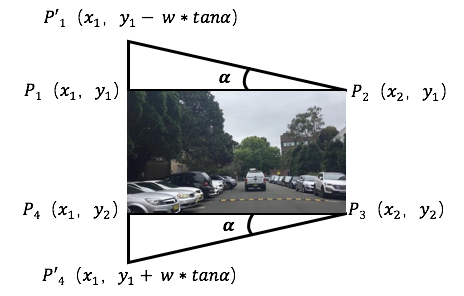}
\caption{\small Skew transform. $P_1, P_2, P_3$ and $P_4$ are original image corner coordinates. $P'_1 and P'_4$ are new points after skew warp. $\alpha$ is the skew magnitude used in this paper ranging from $[10^\circ, 70^\circ]$. $w$ is the image width. }
\label{fig:skew_aug}
\end{center}
\end{figure}

\subsubsection{Skew}

The system proposed in this paper has multiple additional cameras facing in different orientation to cover the entire forward view. Objects captured at the different orientations have explicitly different structure due to changes in perspective. To compensate for this, we apply skew to images to simulate the image structures from side oriented cameras. 

To simulate left camera images, we assume the left side of images are stretched, and same applied to right camera images. As shown in Fig.~\ref{fig:skew_aug}, given the original image plane ($P_1, P_2, P_3, P_4$) and the new stretched plane ($P'_1, P_2, P_3, P'_4$), the homography transform matrix can be obtained by:

\begin{equation}
\begin{bmatrix}
u' \\
v' \\
1
\end{bmatrix}
\sim 
\begin{bmatrix}
h_{11} & h_{12} & h_{13} \\
h_{21} & h_{22} & h_{23}\\
h_{31} & h_{32} & h_{33}
\end{bmatrix}
\begin{bmatrix}
u \\
v \\
1
\end{bmatrix}
\end{equation}
where $(u',v')$ is the new point coordinate, $(u,v)$ is the original point coordinate, and $h_{ii}$ is the parameters in homography matrix. 

Given $h_{33}=1$ and four corresponding points, all parameters $h_{ii}$ can be derived. The skew magnitude is set to be the angle $\alpha$ between original points and new points. In this paper, the skew augmentation is applied with the magnitude angle $\alpha \in [10^\circ, 70^\circ]$ with increments of $10^\circ$ for each training. However, we hold the hypothesis that $\alpha >60^\circ $ will not have great improvement since the objects will be over stretched and become less useful for training.

\subsubsection{Gamma correction}

\begin{figure}[t]
\begin{center}
\includegraphics[width=0.98\columnwidth]{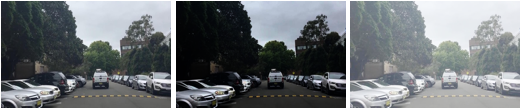}
\caption{\small Gamma correction for Point Grey images. Left to right: original image, image with $\gamma = 0.5$, image with $\gamma = 2.5$. }
\label{fig:gamma}
\end{center}
\end{figure}

Gamma correction, as a photometric transformation, is applied as a data augmentation technique to overcome the variation in illumination between datasets. Ideally, a photon captured by the camera is a linear function. However in real situations, a non-linear power function $ I(x) = L(x)^\gamma$ is always applied where $I(x)$ is the image for display, $L(x)$ is the luminance reaching the camera and $\gamma$ is the correction constant~\cite{gamma}. To change the luminance, we apply the correction equation: $O(x) = I(x)^{1/\gamma}$ so that a smaller $\gamma$ shifts images to lower illumination and larger $\gamma$ results in increased illumination. By varying the value of $\gamma$, the output image $O(x)$ can be changed to simulate under- and over-exposed images (Fig.~\ref{fig:gamma}). 

To choose the best $\gamma$ values and evaluate their influence, we select $\gamma$ for each image from Gaussian distributions truncated between $(0, 3]$. Images with small $\gamma$ values are darker, which emulates the effect of shadows. Images with larger $\gamma$ are brighter and over-exposed. For the Gaussian distribution used to draw sample values of $\gamma$, the mean is set to be $\mu=1$ and the standard deviation $\sigma$ is set within a range between $[0, 1]$ with increments of $0.1$ for each different model.

\begin{table*}
\caption{Gamma correction influence. `Original' is the original dataset without any data augmentation. `ITSC~\cite{itsc_work}' is augmenting original set with flipping, adding noise, blurring and center-cropping. `Gamma $\sigma$' is applying gamma correction from Gaussian distributions with $\mu=1$ and the standard deviations $\sigma \in [0.1, 1]$. Performance is evaluated by per class accuracy (C), mean Intersection over Union (mIoU) and global accuracy (G).}
\label{tab:1}
\definecolor{bluegray}{rgb}{0.4, 0.6, 0.8}
\definecolor{babyblueeyes}{rgb}{0.63, 0.79, 0.95}
\definecolor{beaublue}{rgb}{0.74, 0.83, 0.9}

\begin{center}
\begin{tabular}{|p{2.5cm}|p{2cm}|ccc|ccc|ccc|}

\hline
\multicolumn{2}{|c}{\cellcolor{bluegray}\textbf{Training Sets}}&\multicolumn{6}{|c|}{\cellcolor{bluegray}\textbf{Cityscapes Training Set}} & 
\multicolumn{3}{c|}{\cellcolor{bluegray}\textbf{USYD\_Cloudy\_Set}} \\ \hline
\multicolumn{2}{|c}{\cellcolor{beaublue}\textbf{Validating Sets}}&\multicolumn{3}{|c}{\cellcolor{beaublue}\textbf{Cityscapes Val}} & \multicolumn{3}{|c|}{\cellcolor{beaublue}\textbf{USYD\_Front\_Set}} & \multicolumn{3}{c|}{\cellcolor{beaublue}\textbf{USYD\_Front\_Set}}  \\ \hline 

\multicolumn{2}{|c}{\cellcolor{beaublue}\textbf{Evaluation}}& \multicolumn{1}{|c}{\cellcolor{beaublue}\textbf{C}} & \cellcolor{beaublue}\textbf{mIoU} & \cellcolor{beaublue}\textbf{G} & \cellcolor{beaublue}\textbf{C} & \cellcolor{beaublue}\textbf{mIoU} & \cellcolor{beaublue}\textbf{G} & \cellcolor{beaublue}\textbf{C} & \cellcolor{beaublue}\textbf{mIoU} & \cellcolor{beaublue}\textbf{G}  \\ \hline

\cellcolor{beaublue}&\centering\cellcolor{beaublue}Original&82.94&\textbf{66.78}&\textbf{90.85}&44.20&26.48&61.02&74.49&42.23&81.39  \\ \cline{2-11}
\cellcolor{beaublue}&\centering\cellcolor{beaublue}ITSC~\cite{itsc_work}&86.81&63.54&87.86&55.88&28.34&61.25&75.93&42.93&82.26  \\ \cline{2-11}
\cellcolor{beaublue}&\centering\cellcolor{beaublue}Gamma 0.1&86.30&63.55&88.63&53.66&27.34&65.96&75.40&42.31&81.35 \\ \cline{2-11}
\cellcolor{beaublue}&\centering\cellcolor{beaublue}Gamma 0.2&86.64&66.59&90.01&58.56&30.26&68.00&75.96&43.01&81.65 \\ \cline{2-11}
\cellcolor{beaublue}&\centering\cellcolor{beaublue}Gamma 0.3&\textbf{86.83}&65.46&89.45&52.10&27.26&56.76&\textbf{76.26}&44.13&82.88 \\ \cline{2-11}
\centering\cellcolor{beaublue}\textbf{Augmentation}&\centering\cellcolor{beaublue}Gamma 0.4&86.39&65.07&89.36&54.12&25.56&56.99&75.64&43.78&82.25 \\ \cline{2-11}
\centering\cellcolor{beaublue}\textbf{Method}&\centering\cellcolor{beaublue}Gamma 0.5&86.48&64.38&89.04&54.96&27.18&61.83&76.04&\textbf{45.60}&\textbf{85.03} \\ \cline{2-11}
\cellcolor{beaublue}&\centering\cellcolor{beaublue}Gamma 0.6&86.34&65.96&89.50&59.66&29.65&63.27&75.99&42.27&81.81 \\ \cline{2-11}
\cellcolor{beaublue}&\centering\cellcolor{beaublue}Gamma 0.7&86.53&64.49&88.68&61.02&\textbf{32.60}&\textbf{69.27}&76.05&43.28&82.65 \\ \cline{2-11}
\cellcolor{beaublue}&\centering\cellcolor{beaublue}Gamma 0.8&86.46&65.45&89.29&\textbf{61.77}&31.88&64.46&75.64&43.24&81.91 \\ \cline{2-11}
\cellcolor{beaublue}&\centering\cellcolor{beaublue}Gamma 0.9&86.34&64.96&89.10&57.88&30.14&63.73&75.52&43.94&82.62 \\ \cline{2-11}
\cellcolor{beaublue}&\centering\cellcolor{beaublue}Gamma 1.0&86.62&64.87&89.31&57.68&28.30&62.70&75.99&42.94&81.80 \\ \hline

\end{tabular}
\end{center}
\end{table*}

\begin{figure}[h]
\vspace{1mm}
\begin{center}
\includegraphics[width=0.32\columnwidth]{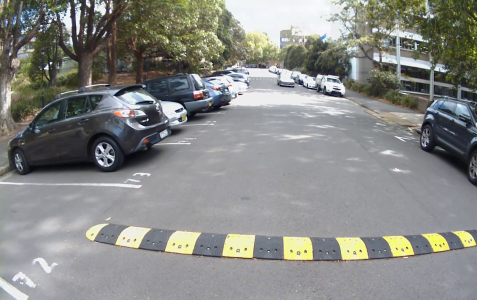}
\includegraphics[width=0.32\columnwidth]{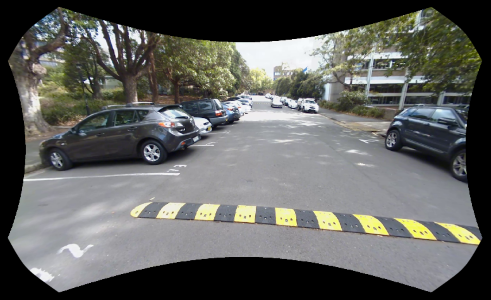}
\includegraphics[width=0.32\columnwidth]{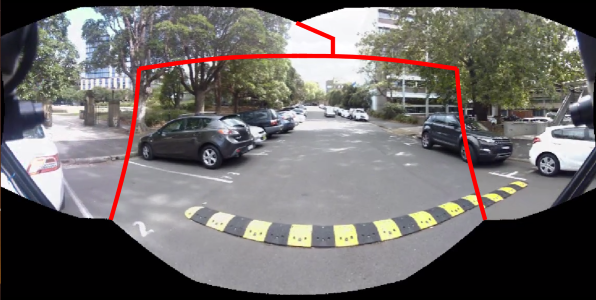}
\caption{\small Demonstration of one frame from the center camera being inserted into the panorama image. The left image is the original image, the center image is after undistortion, and the right image is after the image has been warped and inserted into a panorama canvas with the other two camera images. The edges of these image have been highlighted for clarity.}
\label{fig:cylinder_warping}
\end{center}
\end{figure}

\subsection{Image Stitching}

In this paper, all semantic models are trained on rectified images using five distortion parameters: $k_1, k_2, k_3$ for radial distortion and $p_1, p_2$ for tangential distortion. 

After undistortion, it is common for the images to be cropped as rectangles such that there is no padding visible, however this is undesirable as there is significant information in the corners of these images due to the large distortion of a $100^\circ$ FOV lens. Instead, we perform a full frame undistortion with a binary mask to label the outer image padding. All three of these undistorted images are then aligned using the camera extrinsic parameters and reprojected onto a cylinder, allowing for a full forward facing $180^\circ$ panorama. This differs from a normal `flat' image projection as the `straight line' constraint no longer holds. Instead, vertical straight lines remain straight, but any non-vertical lines are wrapped around the image to prevent stretching of image data very far from its respective optical camera centre ($>$$75^\circ$) and to combine images to create a panorama that exist beyond $180^\circ$.

After each image is warped into the appropriate cylindrical co-ordinates, the images are deposited onto a single master panorama image. This procedure can be seen in Fig. \ref{fig:cylinder_warping}. The image on the left is the original frame captured from the center camera. Notice the significant warping of the speed bump due to the nature of the wide angle lens. The center figure is the image after undistortion. Here padding was used to preserve information in the corner of the frame. The right figure is the result after the image has been warped and deposited with the left and right camera images into a panorama canvas. The edges of these images have been highlighted in red for clarity.

\section{Experiments and Results}
\label{sec:experiment}
In this section we will show quantitative and qualitative results of the semantic segmentation in our local environment. The models were trained only on the original \texttt{USYD\char`_Cloudy\char`_Set} with some data augmentations mentioned earlier. All other datasets were only used for validating the augmentation performance.

\subsection{Training Setup}

As real-time processing is required for autonomous vehicle operation, we adopted the ENet~\cite{enet} architecture for semantic segmentation. The model was trained and tested on a GTX 1080 Ti GPU and also tested on a NVIDIA DRIVE PX2. The network can take arbitrary sized images for both training and testing, and can predict a segmented image with a resolution of $640 \times 360$. The learning rate was set to be $5e-6$ at the beginning and decayed by $1e-1$ when the validation error stopped improving for $100$ epochs. 

Models were firstly trained on the Cityscapes dataset~\cite{cityscapes}, then fine-tuned using our \texttt{USYD\char`_Cloudy\char`_Set}. The original Cityscapes dataset has more than 30 classes, of which a number are not relevant to our local environment. To optimize the network, we remapped these 30 classes into 12 categories to better represent the categories expected in the USYD datasets.  

\begin{figure}[t]
    \centering
\begin{subfigure}{0.9\columnwidth}
  \includegraphics[width=\linewidth]{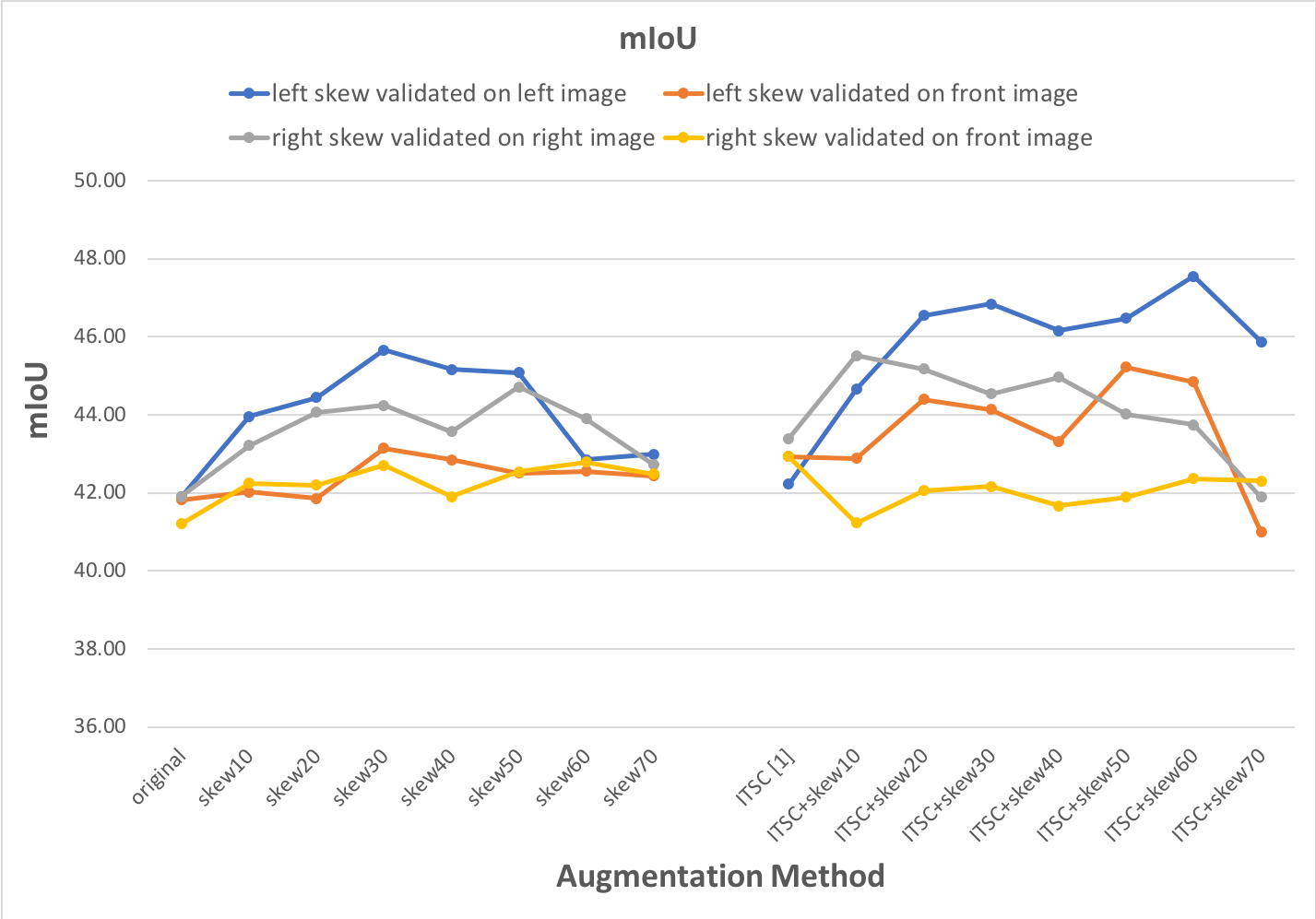}
  \caption{Mean Intersection over Union (mIoU) evaluation.}
  \label{fig:miou}
\end{subfigure}\hfil
\begin{subfigure}{0.9\columnwidth}
  \includegraphics[width=\linewidth]{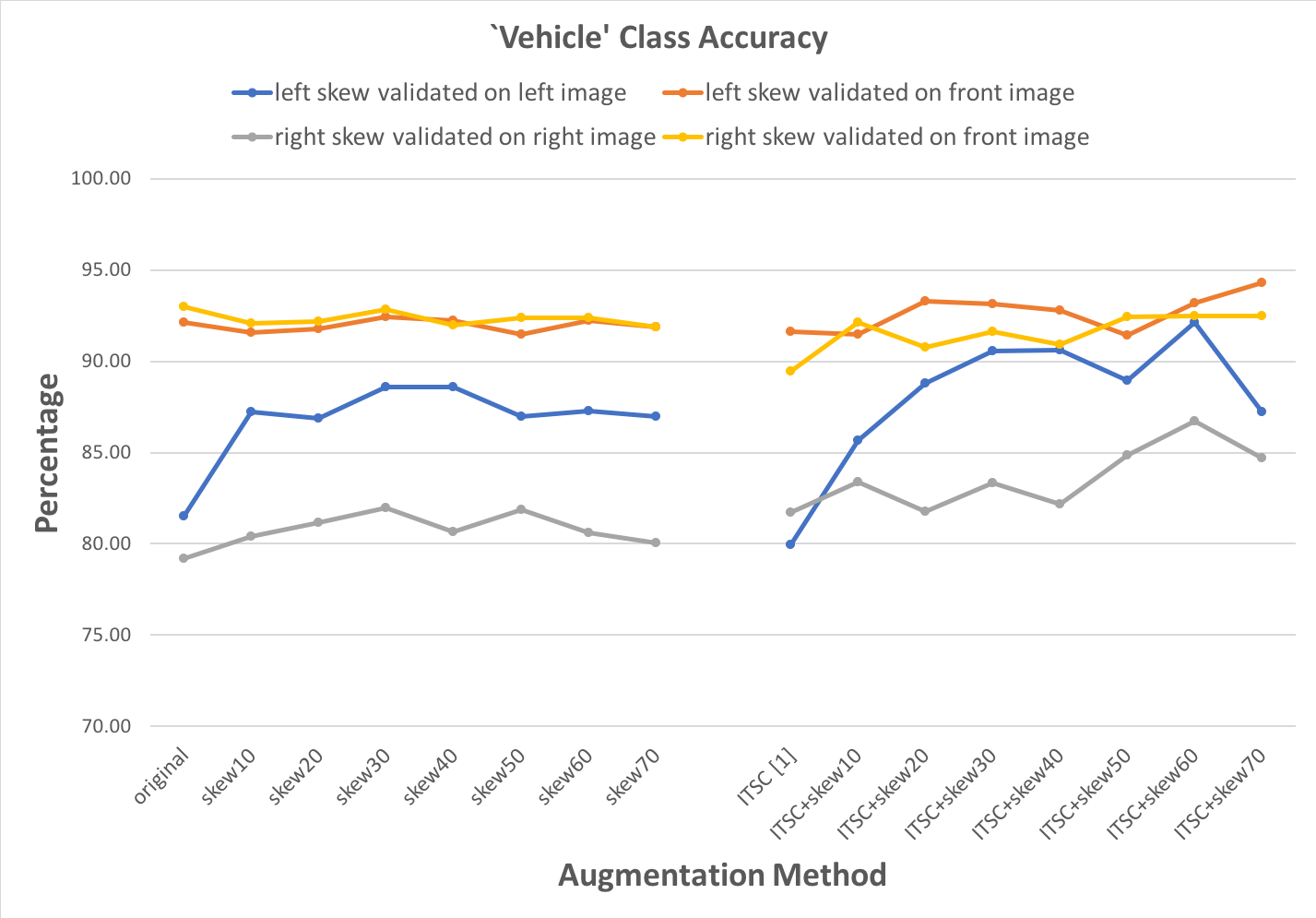}
  \caption{`Vehicle' class accuracy evaluation.}
  \label{fig:vehicle}
\end{subfigure}

\caption{Mean IoU and `Vehicle' class accuracy with skew augmentation tested on front and side-view cameras respectively. The horizontal axis indicates the usage of augmentation methods. `original' is the original \texttt{USYD\char`_Cloudy\char`_Set}. `skewN' is single skew augmentation with magnitude $\alpha=N$. `ITSC' is from previous work with flipping, center-cropping, add noise and blurring.}
\label{fig:skew}
\end{figure}

\begin{figure}[t]
    \centering
\begin{subfigure}{0.95\columnwidth}
  \includegraphics[height=4.75cm,width=\linewidth]{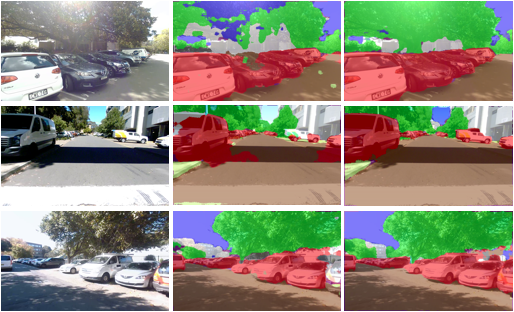}
  \caption{USYD testing results. Row 1: left camera image. Row 2: front camera image. Row 3: right camera image.}
  \label{fig:usyd qua}
\end{subfigure}
~
\begin{subfigure}{0.95\columnwidth}
  \includegraphics[height=5cm,width=\linewidth]{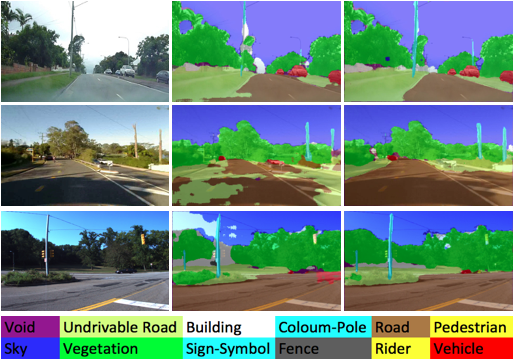}
  \caption{Cross testing results. Row 1: image (resolution: $640 \times 260$) from Alderley, Brisbane, Australia~\cite{day_data}. Row 2: image (resolution: $640 \times 480$) from St Lucia, Brisbane, Australia~\cite{shadow_data}. Row 3: image (resolution: $1024 \times 768$) from CMU Visual Loc.~\cite{cmu}.}
  \label{fig:cross qua}
\end{subfigure}
\caption{Testing results on different datasets. From left to right: original images; segmentation results from \cite{itsc_work} with blurring, flipping, center-cropping and adding noise; results after adding gamma correction and skew.}
\label{fig:qualitative}
\end{figure}

\subsection{Quantitative Analyses}
\subsubsection{Gamma correction}

After looking into the Cityscapes dataset~\cite{cityscapes}, we found that most images in this dataset do not have harsh shadows as with the images we collected in summer. We applied gamma correction to both Cityscapes dataset and  \texttt{USYD\char`_Cloudy\char`_Set} to compare how much the gamma correction can improve the model performance for harsh shadows. The gamma $\gamma$ values were selected from the Gaussian distributions with $\mu=1$ and the standard deviations $\sigma$ from a range of $[0, 1]$ with increments of $0.1$.  

From Table \ref{tab:1}, when the model was trained and validated on Cityscapes dataset, data augmentation only improved per class accuracy while mIoU or global accuracy did not show much improvement. However, when validating on \texttt{USYD\char`_Front\char`_Set} which has strong shadows and over-exposures, gamma correction with $\sigma=0.7$ showed significant improvement. 

After applying gamma correction to \texttt{USYD\char`_Cloudy\char`_Set}, we further fined-tuned the pre-trained model using this dataset. The validation results on \texttt{USYD\char`_Front\char`_Set} showed the most improvement when $\sigma=0.5$, with $1.5\%$ on per class accuracy, $3.5\%$ on mIoU, and $3.6\%$ on global accuracy compared with model fine-tuned on original \texttt{USYD\char`_Cloudy\char`_Set}.

\subsubsection{Skew}

The skew augmentation was applied to simulate the perspective changes of side cameras. Models trained with skew augmentation were validated against images from front cameras and side cameras. 

As mean Intersection over Union (mIoU) shown in Fig.~\ref{fig:miou}, it is clear that models trained with skew augmentations had limited contribution for the forward facing camera images. For the side oriented cameras however, a large improvement is observed (5.7\% with left skew validated on left images and 3.6\% with right skew validated on right images). It also demonstrates that even with single skew augmentation (`skewN' in Fig.~\ref{fig:miou}), the performance surpasses the original `ITSC' model~\cite{itsc_work} which had four random augmentation methods.

The warping from the side oriented cameras is more pronounced on the outer edge of the image (far left for the left camera and far right for the right cameras) as illustrated in Fig. \ref{fig:nvidia}. For much of the dataset, this part of the images contain parked vehicles so we specifically evaluate the performance of the `Vehicle' class to demonstrate the performance. Fig.~\ref{fig:vehicle} clearly shows that prior to the skew augmentation the class accuracy was lower for the side oriented images when compared to front images. After the skew augmentation the performance of the left images are significantly improved, with the performance becoming comparable to the front images. As this vehicle is driven on the left hand side of the road, the left images correspond to the side of the road where the obstacles such as parked cars are closer to the camera. The performance improvement for the right images is less clear, possibly because parked vehicles on the right side are further away and have less perspective change than the left camera.

In general, the skew augmentation magnitude ranging from $20^\circ$ to $50^\circ$ resulted in the highest accuracy improvement. This indicates that smaller magnitudes do not properly represent the changes in perspective, while extremely large magnitudes over scale the objects and are not representative of the real world perspective.

\begin{figure*}[t]
\begin{center}
\vspace{1mm}
\includegraphics[height=3.3cm,width=0.85\textwidth]{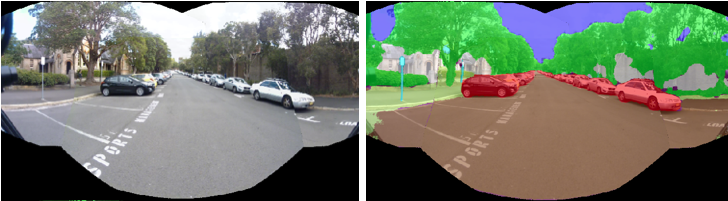}
\caption{\small  $180^\circ$ Semantic segmentation. The first column shows stitching of the original images from left, middle and right cameras (sensor setup is shown in Fig.~\ref{fig:car}). The second column is the semantic segmentation results and path proposal prediction for a $180^\circ$ field of view.}
\label{fig:stitch}
\end{center}
\end{figure*}

\subsection{Qualitative Analyses}

The qualitative improvements from applying data augmentation to the original dataset with gamma correction and skew are illustrated in Fig. \ref{fig:usyd qua}. From the figure, the middle column shows that even with multiple random data augmentation algorithms from the original model \cite{itsc_work}, the over-exposed trees and high contrast shaded areas of road were often incorrectly classified. In addition, the model did not correctly classify the outer portion (closest to the ego vehicle) of the vehicles in the side images due to perspective changes. As demonstrated in the right column of Fig. \ref{fig:usyd qua}, these areas are labeled correctly after applying gamma correction and skew augmentation.

To demonstrate how the proposed improvements transfer to other datasets, we cross-tested the segmentation model with images collected by other groups in different cities. Fig.~\ref{fig:cross qua} shows three examples including an original image, an image that performed poorly due to strong shadows, and a side-view image from each of the three datasets. Adding skew and gamma correction also greatly enhanced the segmentation performance for those images.

\subsection{Image Stitching Results}

After improving the performance of semantic segmentation for the camera array, we are able to stitch the images together to obtain a single combined image covering $180^\circ$ forward facing FOV for the vehicle (shown in Fig. \ref{fig:stitch}). As the raw images are projected to a cylindrical surface without cropping, all pixels are preserved including the corners of the original images.

Adding skew and gamma augmentation enables the CNN models to produce accurate segmentation results for cameras facing different orientations. The stitched images are able to better classify all objects in the front and side of the ego vehicle, and provide a comprehensive scene understanding over the entire forward facing FOV. 

\section{Conclusion}
\label{sec:conclusion}

Semantic segmentation is capable of providing abundant contextual information for autonomous vehicles, though it generally requires a large amount of labeled data to transfer an existing model into a different environment. To avoid expensive hand labeling and obtain a robust understanding of the scene, data augmentation has been widely applied to expand the datasets in many existing works. In this paper we have demonstrated the benefits of specific augmentation algorithms, specifically skew and gamma correction from a practical standpoint to cope with the real-world problems of varying illumination conditions and changing of camera perspectives. By providing quantitative and qualitative analysis for the effect of skew and gamma correction augmentations, we demonstrated that the semantic models can be significantly enhanced within our local environment and also in cross-testing environments.

After improving the robustness of the semantic segmentation for different camera orientations, the stitched image is capable to provide a complete $180^\circ$ forward facing understanding of the surrounding area, which is significant for improving situational awareness in autonomous vehicle operations.

\section*{ACKNOWLEDGMENT}

This work has been funded by the ACFR, Australian Research Council Discovery Grant DP160104081 and University of Michigan / Ford Motors Company Contract `Next Generation Vehicles'.

\bibliographystyle{IEEEtran}
\bibliography{Biblio}

 \newcommand{\noop}[1]{}
\begin{thebibliography}{10}
\providecommand{\url}[1]{#1}
\csname url@samestyle\endcsname
\providecommand{\newblock}{\relax}
\providecommand{\bibinfo}[2]{#2}
\providecommand{\BIBentrySTDinterwordspacing}{\spaceskip=0pt\relax}
\providecommand{\BIBentryALTinterwordstretchfactor}{4}
\providecommand{\BIBentryALTinterwordspacing}{\spaceskip=\fontdimen2\font plus
\BIBentryALTinterwordstretchfactor\fontdimen3\font minus
  \fontdimen4\font\relax}
\providecommand{\BIBforeignlanguage}[2]{{%
\expandafter\ifx\csname l@#1\endcsname\relax
\typeout{** WARNING: IEEEtran.bst: No hyphenation pattern has been}%
\typeout{** loaded for the language `#1'. Using the pattern for}%
\typeout{** the default language instead.}%
\else
\language=\csname l@#1\endcsname
\fi
#2}}
\providecommand{\BIBdecl}{\relax}
\BIBdecl

\bibitem{itsc_work}
W.~Zhou, R.~Arroyo, A.~Zyner, J.~Ward, S.~Worrall, E.~Nebot, and L.~M. Bergasa,
  ``Transferring visual knowledge for a robust road environment perception in
  intelligent vehicles,'' in \emph{IEEE 20th International Conference on
  Intelligent Transportation Systems (ITSC)}, 2017, pp. 1483--1488.

\bibitem{cityscapes}
M.~Cordts, M.~Omran, S.~Ramos, T.~Rehfeld, M.~Enzweiler, R.~Benenson,
  U.~Franke, S.~Roth, and B.~Schiele, ``The cityscapes dataset for semantic
  urban scene understanding,'' in \emph{Proceedings of the IEEE Conference on
  Computer Vision and Pattern Recognition}, 2016, pp. 3213--3223.

\bibitem{autoaugment}
E.~D. Cubuk, B.~Zoph, D.~Mane, V.~Vasudevan, and Q.~V. Le, ``Autoaugment:
  Learning augmentation policies from data,'' \emph{arXiv preprint
  arXiv:1805.09501}, 2018.

\bibitem{augmentation_jesus}
J.~Munoz-Bulnes, C.~Fernandez, I.~Parra, D.~Fern{\'a}ndez-Llorca, and M.~A.
  Sotelo, ``Deep fully convolutional networks with random data augmentation for
  enhanced generalization in road detection,'' in \emph{Intelligent
  Transportation Systems (ITSC), 2017 IEEE 20th International Conference
  on}.\hskip 1em plus 0.5em minus 0.4em\relax IEEE, 2017, pp. 366--371.

\bibitem{kitti}
J.~Fritsch, T.~Kuhnl, and A.~Geiger, ``A new performance measure and evaluation
  benchmark for road detection algorithms,'' in \emph{Intelligent
  Transportation Systems-(ITSC), 2013 16th International IEEE Conference
  on}.\hskip 1em plus 0.5em minus 0.4em\relax IEEE, 2013, pp. 1693--1700.

\bibitem{fcn}
J.~Long, E.~Shelhamer, and T.~Darrell, ``Fully convolutional networks for
  semantic segmentation,'' in \emph{Proceedings of the IEEE Conference on
  Computer Vision and Pattern Recognition}, 2015, pp. 3431--3440.

\bibitem{refinenet}
G.~Lin, A.~Milan, C.~Shen, and I.~Reid, ``Refinenet: Multi-path refinement
  networks with identity mappings for high-resolution semantic segmentation,''
  \emph{arXiv preprint arXiv:1611.06612}, 2016.

\bibitem{deeplabv2}
L.-C. Chen, G.~Papandreou, I.~Kokkinos, K.~Murphy, and A.~L. Yuille, ``Deeplab:
  Semantic image segmentation with deep convolutional nets, atrous convolution,
  and fully connected crfs,'' \emph{arXiv preprint arXiv:1606.00915}, 2016.

\bibitem{pspnet}
H.~Zhao, J.~Shi, X.~Qi, X.~Wang, and J.~Jia, ``Pyramid scene parsing network,''
  \emph{arXiv preprint arXiv:1612.01105}, 2016.

\bibitem{deeplabv3}
L.-C. Chen, G.~Papandreou, F.~Schroff, and H.~Adam, ``Rethinking atrous
  convolution for semantic image segmentation,'' \emph{arXiv preprint
  arXiv:1706.05587}, 2017.

\bibitem{segnet}
V.~Badrinarayanan, A.~Kendall, and R.~Cipolla, ``Segnet: A deep convolutional
  encoder-decoder architecture for image segmentation,'' \emph{IEEE
  Transactions on Pattern Analysis and Machine Intelligence}, 2017.

\bibitem{enet}
A.~Paszke, A.~Chaurasia, S.~Kim, and E.~Culurciello, ``Enet: A deep neural
  network architecture for real-time semantic segmentation,'' \emph{arXiv
  preprint arXiv:1606.02147}, 2016.

\bibitem{fisheye1}
L.~Deng, M.~Yang, Y.~Qian, C.~Wang, and B.~Wang, ``Cnn based semantic
  segmentation for urban traffic scenes using fisheye camera,'' in
  \emph{Intelligent Vehicles Symposium (IV), 2017 IEEE}.\hskip 1em plus 0.5em
  minus 0.4em\relax IEEE, 2017, pp. 231--236.

\bibitem{supersensor}
R.~Varga, A.~Costea, H.~Florea, I.~Giosan, and S.~Nedevschi, ``Super-sensor for
  360-degree environment perception: Point cloud segmentation using image
  features,'' in \emph{IEEE 20th International Conference on Intelligent
  Transportation Systems (ITSC)}, 2017, pp. 126--132.

\bibitem{imagenet}
O.~Russakovsky, J.~Deng, H.~Su, J.~Krause, S.~Satheesh, S.~Ma, Z.~Huang,
  A.~Karpathy, A.~Khosla, M.~Bernstein \emph{et~al.}, ``Imagenet large scale
  visual recognition challenge,'' \emph{International Journal of Computer
  Vision}, vol. 115, no.~3, pp. 211--252, 2015.

\bibitem{alexa}
A.~Carlson, K.~A. Skinner, and M.~Johnson-Roberson, ``Modeling camera effects
  to improve deep vision for real and synthetic data,'' \emph{arXiv preprint
  arXiv:1803.07721}, 2018.

\bibitem{aug1}
K.~Chatfield, K.~Simonyan, A.~Vedaldi, and A.~Zisserman, ``Return of the devil
  in the details: Delving deep into convolutional nets,'' \emph{arXiv preprint
  arXiv:1405.3531}, 2014.

\bibitem{aug2}
J.~Lemley, S.~Bazrafkan, and P.~Corcoran, ``Smart augmentation-learning an
  optimal data augmentation strategy,'' \emph{IEEE Access}, 2017.

\bibitem{labelme}
A.~Torralba, B.~C. Russell, and J.~Yuen, ``Labelme: Online image annotation and
  applications,'' \emph{Proceedings of the IEEE}, vol.~98, no.~8, pp.
  1467--1484, 2010.

\bibitem{day_data}
M.~J. Milford and G.~F. Wyeth, ``Seqslam: Visual route-based navigation for
  sunny summer days and stormy winter nights,'' in \emph{Robotics and
  Automation (ICRA), 2012 IEEE International Conference on}.\hskip 1em plus
  0.5em minus 0.4em\relax IEEE, 2012, pp. 1643--1649.

\bibitem{shadow_data}
A.~J. Glover, W.~P. Maddern, M.~J. Milford, and G.~F. Wyeth, ``Fab-map+
  ratslam: Appearance-based slam for multiple times of day,'' in \emph{Robotics
  and Automation (ICRA), 2010 IEEE International Conference on}.\hskip 1em plus
  0.5em minus 0.4em\relax IEEE, 2010, pp. 3507--3512.

\bibitem{cmu}
\BIBentryALTinterwordspacing
H.~Badino, D.~Huber, and T.~Kanade. (2011) The cmu visual localization data
  set. [Online]. Available:
  \url{http://3dvis.ri.cmu.edu/data-sets/localization}
\BIBentrySTDinterwordspacing

\bibitem{gamma}
A.~Galdran, A.~Alvarez-Gila, M.~I. Meyer, C.~L. Saratxaga, T.~Ara{\'u}jo,
  E.~Garrote, G.~Aresta, P.~Costa, A.~M. Mendon{\c{c}}a, and A.~Campilho,
  ``Data-driven color augmentation techniques for deep skin image analysis,''
  \emph{arXiv preprint arXiv:1703.03702}, 2017.

\end{thebibliography}

\end{document}